\documentclass[10pt,twocolumn,letterpaper]{article}
\usepackage[pagenumbers]{wacv}

\usepackage{microtype}
\usepackage{longtable}
\usepackage{array}
\usepackage{seqsplit}
\newcommand{\ev}[1]{\textsuperscript{\textsf{[#1]}}}
\newcommand{\pp}{\,\mathrm{pp}}

\definecolor{wacvblue}{rgb}{0.21,0.49,0.74}
\usepackage[breaklinks,colorlinks,allcolors=wacvblue]{hyperref}

\title{Answer Correction with Verifiable Rewards:\\A Gemma-Family Study of Multi-View Driving QA}

\author{T N Rickesh$^1$ \qquad Pranav Vinodh$^2$ \qquad Sowmya Kamath S.$^2$ \\
$^1$natkal \qquad $^2$National Institute of Technology Karnataka \\
{\tt\small info@natkal.ai \qquad prnv.231it050@nitk.edu.in \qquad sowmyakamath@nitk.edu.in}
}

\begin{document}
\maketitle

\begin{abstract}
Multi-camera driving question answering provides a discrete test of vision-language models, but post-training gains can reflect changed selected answers or improved answer extraction. We investigate this distinction within the Gemma model family using archived, task-linked WaymoQA evaluations. A zero-shot diagnostic on 5,519 test questions spans effective-small, mixture-of-experts, unified, and dense checkpoints, with serving differences disclosed rather than interpreted as architecture effects. The strongest controlled evidence is a matched-harness Gemma-4-12B base/GRPO pair: accuracy rises from 73.78\% to 75.43\%, a gain of 1.65 percentage points with a scene-cluster bootstrap interval of [+0.84, +2.41]. There are 215 improvements and 124 regressions; 213 improvements start from an extractable but incorrect base answer, and only two recover previously unextractable responses. Crossing correctness with extraction state shows why correct noncompliant outputs cannot be counted as prediction errors. Category counts reveal a critical safety asymmetry: routine perception gains (+5.93 pp) mask a regression in safety-critical hazard prediction ($-1.82$ pp), demonstrating that unweighted outcome rewards can trade off rare edge cases for aggregate accuracy. Reward saturation describes sampled training dynamics rather than an architectural ceiling. A separate archived Gemma-31B thinking comparison yields no detected average improvement, with uncertainty that does not establish equivalence. These results support a modest answer-selection gain for the recorded tuning configuration and a reproducible diagnostic accounting, without attributing corrected labels to specific grounding mechanisms or claiming superiority over matched supervised tuning.
\end{abstract}

\section{Introduction}
In autonomous driving world action models, agents must determine safe action policies from panoramic multi-camera observations. Multi-view driving question answering evaluates whether foundation models can select valid navigation actions under structured prompts before integration into closed-loop control. General-purpose multimodal models can perform this task without driving-specific training, making their zero-shot behavior a useful starting point for adaptation. WaymoQA supplies multi-view questions about perception, prediction, planning, and safety-related scenarios~\citep{waymoqa}. Its discrete answer labels make verifiable-reward tuning feasible, but also make it easy to mistake a better scored answer for evidence of better visual grounding.

This paper studies Gemma-family checkpoints and asks a narrower question: \emph{does the observed post-training gain mainly correct extracted wrong options, or make previously unavailable answers extractable?} These are different changes in an evaluation pipeline. A model may violate an answer-only JSON instruction while still selecting the correct option. Conversely, a strictly formatted wrong answer is not evidence that its underlying error was caused by camera seams, coordinate projection, or counterfactual branching. Establishing those causes requires visual interventions or annotated evidence.

The Gemma family is relevant because it spans distinct multimodal designs rather than one fixed architecture. We first describe its zero-shot diagnostic as a serving-conditional reference. We then center the paper on a base/adapter Gemma-4-12B comparison using the same evaluator and test membership. Its net gain is modest, but almost every wrong-to-correct transition begins with an extractable base response. This is evidence about answer selection under the recorded parser; it is not a reward-component ablation or a causal grounding diagnosis.

Our contributions are a family-specific multi-view diagnostic, a paired transition analysis of verifiable-reward tuning, and an outcome accounting that keeps correctness separate from compliance. Supporting analyses report category counts, sampled training dynamics, and uncertainty for an archived thinking configuration. All numerical results map to task evaluation artifacts or explicitly labeled reconstructions. The study contains no new training algorithm and no matched SFT control.

\paragraph{Relationship to the companion paper.}
A companion study examines finite-budget answer delivery in an Ornith-style reasoning-enabled derivative~\citep{companionornith}. The two manuscripts reuse the WaymoQA board and selected reference tasks. This paper's training result is the Gemma-12B pair; the companion's result concerns Ornith generation, extraction, and recovery. We disclose shared provenance rather than count the common benchmark as an independent contribution. The checkpoint-family split is not a claim that all general-purpose and derivative models occupy disjoint architecture classes.

\section{Related Work and Background}

\subsection{Gemma-Style Multi-Modal Architectures}
We use \emph{Gemma-style} here for the Gemma-family multimodal checkpoints in the task records. Google's documentation distinguishes effective-small models with per-layer embeddings, a dense model, a mixture-of-experts model, and an encoder-free unified model using direct input projections~\citep{gemmadocs}. Our evaluated names are Gemma-4-E4B, Gemma-4-26B-A4B, Gemma-4-12B, and Gemma-4-31B. ``Unified'' identifies the documented 12B design; it is not a property we assign identically to every checkpoint. The family also supports configurable thinking~\citep{gemmadocs}. Our principal 12B evaluation suppresses that channel and requests a direct answer; direct answering is an experimental setting, not evidence that Gemma models lack reasoning. We do not attribute differences among family rows to encoder design, routing, or per-layer embeddings because the serving stacks and generation caps are not controlled across those rows.

\subsection{World Action Models and Multimodal Driving Policies}
World models in autonomous driving aim to forecast scene dynamics, simulate counterfactual futures, and determine vehicle action policies conditioned on panoramic sensor inputs. GAIA-1~\citep{hu2023gaia1} introduces a generative world model using video, text, and action tokens to predict future driving states. UniAD~\citep{hu2023uniad} and VAD~\citep{jiang2023vad} structure autonomous driving around perception, prediction, and planning representations. Copilot4D~\citep{zhang2024copilot4d} learns unsupervised world models via discrete diffusion over dynamic sensor sequences.

While closed-loop simulators evaluate driving trajectories over time, open-loop benchmarks test whether foundation models can accurately perceive visual affordances and select correct navigation actions. DriveLM~\citep{drivelm} formats driving perception and action selection as visual question answering over scene graphs. WaymoQA~\citep{waymoqa} extends this paradigm to synchronized multi-view observations across eight cameras, providing discrete queries spanning object tracking, trajectory prediction, and safety-critical intervention decisions. Surveys by Cui et al.~\citep{cui2023survey} emphasize that deploying multimodal foundation models to autonomous driving requires evaluating both reasoning accuracy and structural safety constraints. We study these multi-view questions as an action selection benchmark for evaluating general-purpose foundation models.

\subsection{Reinforcement Learning with Verifiable Rewards in Embodied AI}
Reinforcement learning with verifiable rewards (RLVR) trains language and multimodal policies using automated rule-based verifiers rather than learned reward models. Shao et al.~\citep{deepseekmath} introduced Group Relative Policy Optimization (GRPO), computing policy gradients from relative rewards across sampled rollouts. DeepSeek-R1~\citep{deepseekr1} demonstrated that binary rule-based rewards incentivize reasoning in mathematical domains.

Extending RLVR to embodied agents introduces distinct dynamics. Carta et al.~\citep{carta2023grounding} evaluate reinforcement learning for grounding language models in interactive environments, finding that reward shaping is sensitive to environment complexity. In autonomous driving, Think2Drive~\citep{li2024think2drive} trains driving agents by planning within latent world models using reinforcement learning. In our visual task, the verifier checks the supplied option label and output contract. It does not check which images support the answer. That distinction motivates the paired transition analysis and prevents us from treating outcome agreement as a verified chain of visual reasoning.

\subsection{Action Selection versus Syntax Adherence in Structured Outputs}
Deploying vision-language models into robotic and vehicle architectures requires structured communication, typically enforced through JSON schemas or special delimiters. However, format compliance and decision accuracy represent orthogonal axes of model behavior. Mishra et al.~\citep{mishra2022cross} and Wang et al.~\citep{wang2023camels} show that instruction tuning often alters surface-level syntax without improving semantic reasoning.

In safety-critical driving contexts, this distinction is consequential. An agent that emits valid JSON selecting an unsafe lane change remains defective. Disentangling whether post-training gains stem from correcting substantive action errors or recovering non-compliant outputs is necessary before attributing improvements to enhanced visual scene understanding.

\section{Dataset and Recorded Evaluation Protocol}
\subsection{Test membership and reconstructed training pool}
The test board contains 5,519 questions from 54 scene tokens. Its JSONL content hash matches the dataset task state, and its rows identify their source split as test.\ev{E4} The two Gemma-12B evaluation tasks use this dataset identifier. We validate every archived item index, category, and answer label against the board.\ev{E1--E2}

The Gemma training task logs 12,297 selected examples with eight views.\ev{E3} A recovered local training file matches that count and the logged answer histogram, spans 148 scene tokens, and has no scene-token intersection with the test board. Historical training-file bytes were not content-hashed by the task, so this is a \emph{reconstructed} partition check. Matching counts and labels is not byte authentication. We report the training pool, not an assertion that every pool example was visited during the recorded optimizer steps. The historical checkpoint-selection boundary is also insufficiently documented to certify an untouched selection process.

\subsection{Harness disclosure}
The matched 12B pair uses the same Hugging Face evaluator revision and task model identifier \texttt{google/gemma-4-12b-it}. Both conditions use NF4 quantization, greedy decoding, a 256-token generation cap, and images resized with aspect ratio preserved to a maximum dimension of 768 pixels.\ev{E1--E2} The prompt requests one answer JSON object, supplies an example, and suppresses the thought channel through the chat template.

Recorded view order is front-left, front, front-right, side-right, side-left, rear-left, rear, rear-right.\ev{E1--E3} The historical evaluator can substitute a black image for a missing view, but the output artifacts do not include a per-item replacement flag. A contemporaneous frame-availability audit remains necessary before diagnosing residual visual errors. We do not assume an identical visual-token count for every model or describe aspect-preserving resizing as a square-image transform.

The zero-shot ladder uses additional services. E4B and 26B tasks record Ollama; the 31B task records an API-compatible backend routed through a local endpoint.\ev{E13--E15} A backend label alone does not prove endpoint implementation or model precision. We therefore leave those precisions uncertified and use these rows only as descriptive references. Cross-service rankings are not paired architecture or training-effect estimates.

\subsection{Metrics and error accounting}
The archived extractor accepts strict answer JSON and several fallback forms, including answer tags and explicit prose answers. A stricter check removes recognized thought-channel markers and then requires the answer JSON alone. We reproduce both checks from the evaluator snapshot and verify the saved item flags.\ev{E1--E2}

Let $C_i$ indicate that the extracted option equals the ground-truth label, $Q_i$ indicate that an option is extractable, and $F_i$ indicate strict post-sanitization compliance. Accuracy is $N^{-1}\sum_i C_i$ over all requested items. An unavailable output contributes zero. Cross these indicators to obtain five mutually exclusive outcomes: strict/correct, strict/wrong, fallback/correct, fallback/wrong, and unextractable. A format violation can coincide with correctness; it cannot be allocated wholesale to prediction error.

For paired predictions, we count wrong-to-correct and correct-to-wrong transitions, and partition improvements according to the \emph{base} extraction state. We report percentile 95\% scene-cluster intervals using 10,000 resamples and analysis seed 42; each replicate samples scenes with replacement and recomputes the pooled item-weighted difference. Exact item-level McNemar tests are secondary because questions sharing a scene are dependent. Neither procedure accounts for training-seed variation, repeated experiment inspection, or undocumented checkpoint selection.

\section{Zero-Shot Diagnostic within the Gemma Family}
\begin{table}[t]
\centering\small\resizebox{\columnwidth}{!}{\begin{tabular}{llrrr}
\toprule
Checkpoint & Recorded service & Correct & Answered & Acc. all (\%) \\
\midrule
Gemma-4-E4B & Ollama & 2,998 & 5,519 & 54.32 \ev{E15} \\
Gemma-4-26B-A4B & Ollama & 3,743 & 5,519 & 67.82 \ev{E14} \\
Gemma-4-12B & HF / NF4 & 4,072 & 5,515 & 73.78 \ev{E1} \\
Gemma-4-31B & API-compatible endpoint & 4,261 & 5,518 & 77.21 \ev{E13} \\
\bottomrule
\end{tabular}
}
\caption{Recorded zero-shot Gemma-family results on 5,519 requested items. ``Answered'' denotes an extracted prediction in these artifacts. Precision is recorded as NF4 for the matched 12B evaluator; it is not authenticated for the other services. The ladder is descriptive and does not isolate model size or architecture.}
\label{tab:zs}
\end{table}

The full-denominator ladder ranges from 54.32\% for E4B to 77.21\% for 31B.\ev{E13--E15} The latter has 4,261 correct labels and 5,518 extracted predictions; its answered-item accuracy is 77.22\%. That conditional score must not replace its all-item score. The 12B base gives 4,072 correct labels or 73.78\% under its separate recorded harness.\ev{E1}

These rows motivate adaptation of a model that already answers most questions in a recoverable form. They do not show that a particular Gemma architecture causes superior grounding. Nor do they establish superiority to published supervised results obtained under another harness. Such comparisons are intentionally absent from the title and primary claim. Our family-level diagnostic is a reference for the matched 12B experiment, not a substitute for its controls.

\section{Verifiable-Reward Tuning and Paired Results}
\subsection{Recorded training configuration}
The 12B task specifies QLoRA rank 64 and scaling 64 on attention projections plus gate/up/down feed-forward projections, eight sampled completions, temperature 0.7, a 256-token completion limit, learning rate $10^{-5}$, and 500 optimizer steps.\ev{E3} Its objective optimizes a composite verifiable reward $R(y, y^*) = R_{\text{correct}}(y, y^*) + R_{\text{format}}(y) + R_{\text{extract}}(y)$, where $y$ is the sampled completion, $y^*$ is the target label, and $\mathcal{E}(y)$ denotes the extracted option:
\begin{equation}
R_{\text{correct}}(y, y^*) = \begin{cases} 
+2.0, & \text{if } \mathcal{E}(y) = y^* \\ 
-1.0, & \text{if } \mathcal{E}(y) \neq y^* \land \mathcal{E}(y) \neq \emptyset \\ 
-2.0, & \text{if } \mathcal{E}(y) = \emptyset,
\end{cases}
\end{equation}
with strict format reward $R_{\text{format}}(y) = \pm 0.5$ and extractability reward $R_{\text{extract}}(y) = \pm 0.25$. These values come from the task-pinned trainer and reward contract; no additional reward weights are inferred.

The objective checks the selected answer and its form, not its visual rationale. Combining reward terms also prevents a causal attribution of any gain to one component. The related 31B training task records 350 steps; its separate serving comparison is not used to estimate a GRPO effect.\ev{E9}

\subsection{Action Policy Correction Decoupled from Syntax Compliance}
The base and adapter score 4,072 and 4,163 correct answers on the same 5,519 test items: 73.78\% and 75.43\%.\ev{E1--E2} The net difference is 91 correct answers, or $+1.65\pp$, with scene-cluster interval $[+0.84,+2.41]\pp$. There are 215 improvements and 124 regressions; item-level McNemar $p=8.82\times10^{-7}$. Both are correct on 3,948 questions and wrong on 1,232. The interval characterizes this paired evaluation, not variability across independently trained adapters.

\begin{table}[t]
\centering\small\begin{tabular}{lrr}
\toprule
Outcome & Base & GRPO \\
\midrule
Strict, correct & 3,767 & 3,922 \\
Strict, wrong & 1,364 & 1,297 \\
Fallback, correct & 305 & 241 \\
Fallback, wrong & 79 & 59 \\
Unextractable & 4 & 0 \\
\midrule
Total & 5,519 & 5,519 \\
\bottomrule
\end{tabular}

\caption{Gemma-12B correctness crossed with archived extraction state.\ev{E1--E2} Correct fallback answers remain correct, despite violating the requested format. The five outcomes sum to the full denominator.}
\label{tab:joint}
\end{table}

\begin{figure}[t]
\centering\includegraphics[width=\linewidth]{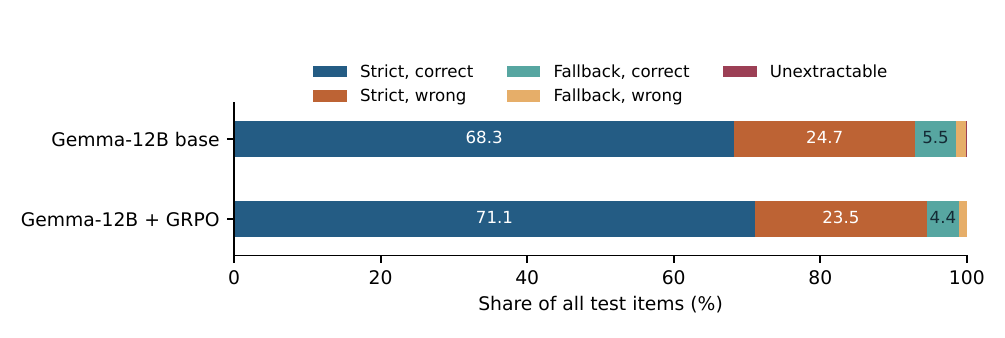}
\caption{Observed response outcomes for the matched Gemma pair.\ev{E1--E2} Accuracy and strict compliance measure overlapping properties. Exact small-segment counts are in Table~\ref{tab:joint}; no segment diagnoses a visual failure mechanism.}
\label{fig:joint}
\end{figure}

The base produces 388 non-strict responses, of which 305 score correctly through fallback extraction, or 78.61\%. Its prediction errors consist of 1,443 extractable wrong answers and four unextractable responses.\ev{E1} Figure~\ref{fig:joint} visualizes these operational states rather than an inferred convention/capability partition.

Among 215 improvements, 213 begin with an extractable wrong base option and two with an unextractable response.\ev{E1--E2} Thus recovered extraction contributes at most $2/5{,}519=0.036\pp$ to the gross positive transitions. This bounds the observed answer-recovery contribution under the archived parser. It demonstrates that verifiable-reward tuning directly shifts the action selection policy across driving queries, rather than acting as a superficial syntax repair operator.

Strict compliance rises from 92.97\% to 94.56\%, while extractability rises from 99.93\% to 100\%.\ev{E1--E2} These changes accompany answer correction but should not be substituted for it. Equally, the small number of unavailable base answers leaves little scope for extraction alone to explain the net accuracy gain.

\subsection{Safety Asymmetry: Routine Perception Gains versus Safety-Critical Prediction Regressions}
The aggregate improvement is not uniform across categories, as shown in Table~\ref{tab:categories}.

\begin{table}[t]
\centering\small\resizebox{\columnwidth}{!}{\begin{tabular}{lrrrr}
\toprule
Category & $n$ & Base correct & GRPO correct & $\Delta$ (pp) \\
\midrule
Counterfactual & 179 & 106 & 111 & +2.79 \\
Normal Perception \& Prediction & 691 & 459 & 500 & +5.93 \\
Normal Planning & 258 & 190 & 193 & +1.16 \\
Normal scene Understanding & 768 & 532 & 529 & -0.39 \\
Normal scene caption & 135 & 106 & 111 & +3.70 \\
Safety-critical Perception & 253 & 165 & 167 & +0.79 \\
Safety-critical Planning & 607 & 448 & 465 & +2.80 \\
Safety-critical Prediction & 659 & 542 & 530 & -1.82 \\
Safety-critical scene Understanding & 281 & 206 & 211 & +1.78 \\
Safety-critical scene caption & 158 & 91 & 96 & +3.16 \\
Special Event & 841 & 737 & 745 & +0.95 \\
Traffic Signal & 244 & 196 & 194 & -0.82 \\
Uncertainty & 445 & 294 & 311 & +3.82 \\
\bottomrule
\end{tabular}
}
\caption{Gemma-12B performance across WaymoQA categories on 5,519 test items.\ev{E1--E2} Net gains cluster in routine perception, while safety-critical prediction exhibits a regression.}
\label{tab:categories}
\end{table}

In standard categories, the model exhibits notable improvements: Normal Perception \& Prediction increases by +5.93 percentage points (459 to 500 correct, +41 net), Normal Planning increases by +1.16 percentage points, and Counterfactual Reasoning increases by +2.79 percentage points. 

In contrast, in Safety-critical Prediction (predicting hazard trajectories, pedestrian crossing intents, and emergency interventions), performance regresses by $-1.82$ percentage points, falling from 542 to 530 correct answers ($-12$ net). 

This divergence reveals an important property of outcome-based post-training in driving world action models: optimizing an unweighted binary reward objective over a mixed task distribution allows the policy to trade off accuracy on difficult safety-critical edge cases in exchange for higher success rates on routine perception queries.

\subsection{Reward Variance Collapse and Gradient Starvation}
The recorded training logs provide insight into the limits of discrete outcome rewards in driving tasks. Over the 500 optimizer steps, the logged group reward increases from $-2.75$ at initialization to $+2.75$ at convergence.\ev{E3} 

However, as training progresses, the within-group reward standard deviation collapses to zero, and the fraction of zero-variance groups reaches 1.00 (\texttt{frac\_reward\_zero\_std = 1.00}), as illustrated in Figure~\ref{fig:training}. Because GRPO relies on group-relative advantage normalization, when all eight rollouts within a prompt group produce identical answers (all correct on easy scenes, or all incorrect on challenging edge cases), the group-relative advantage vanishes, producing zero gradient updates for those items. 

This diagnostic indicates that discrete binary rewards in driving world models suffer from gradient starvation once basic tasks are mastered. Sustaining policy improvements on rare safety-critical scenarios requires continuous reward signals or cost-sensitive weighting:
\begin{equation}
R_{\text{safe}}(y, y^*, c) = w_c \cdot R_{\text{correct}}(y, y^*) + R_{\text{format}}(y) + R_{\text{extract}}(y),
\end{equation}
where category weight $w_c > 1.0$ for safety-critical hazard queries penalizes safety false predictions during policy gradient updates.

\subsection{Thinking on an archived Gemma-31B configuration}
Direct answering gives 4,334 correct labels out of 5,519 (78.53\%); the thinking journals give 4,320 (78.28\% over all requested items).\ev{E8} Thinking has 5,491 answered items, producing 78.67\% conditional answered-item accuracy. Keeping these denominators separate prevents a failed-output exclusion from becoming an apparent benefit.

There are 294 thinking improvements and 308 regressions. Thinking minus direct accuracy is $-0.25\pp$, with scene interval $[-1.18,+0.68]\pp$ and item-level McNemar $p=0.596$.\ev{E8} This is no detected average improvement in the archived configurations. It is not equivalence: no prospective equivalence margin was set, and the interval spans practically relevant differences. Unconstrained test-time thinking without reward grounding produces no detected benefit on multi-view driving question answering, demonstrating that additional token generation does not automatically resolve panoramic spatial ambiguities.

\section{Scope, Limitations, and Conclusion}
The principal result is a modest matched-harness answer-selection gain for the recorded 12B adapter. It does not establish that GRPO is necessary, better than SFT, or uniquely responsible for improved grounding. A matched supervised answer-only baseline and reward-component ablations are missing. Complete historical training-byte and checkpoint-selection authentication is also missing, despite a reproducible no-overlap check on the recovered training file.

Wrong extracted options may arise from many mechanisms, including answer priors or imperfect image processing. Without a blinded annotation audit and controlled input ablations, we cannot assign cross-camera identity loss, coordinate errors, overconfidence, or counterfactual collapse to those rows. The historical black-image fallback strengthens the need for a frame audit before making such diagnoses. These benchmark results also do not certify closed-loop driving safety.

The Gemma-family study nevertheless establishes the operational finding that makes this paper distinct: most observed improvements correct an already extracted wrong option, while strict compliance is a separate measurement. The zero-shot ladder supplies family context, the paired experiment supplies the primary post-training result, and the category and training traces delimit its interpretation. A companion manuscript separately studies answer availability and completion budgets in Ornith-style reasoning-enabled derivatives~\citep{companionornith}.

{
    \small
    \bibliographystyle{ieeenat_fullname}
    \bibliography{references_wamad}
}

\appendix
\section{Provenance and Reproduction}
Cached artifacts and full checksums are retained in the shared author audit directory, with no model inference required to reproduce the statistics. The split package includes a numerical ledger and scripts to regenerate the manuscript tables and figures from that cache.

The authenticated board SHA-256 is:
\begin{flushleft}\small\texttt{\seqsplit{c170357880de6ab2c35b8072714ee8422a5de3cb6173fbd8849b9fdbe045e620}}\end{flushleft}
The recovered training file's current SHA-256 is:
\begin{flushleft}\small\texttt{\seqsplit{7fbcfde610e1a75c24fbce012055f58343a446dbcbffcbfa2b4d2e59a77c0780}}.\end{flushleft}
Its label counts are A: 3,440; B: 2,980; C: 2,870; D: 3,007, matching the training log.\ev{E3} This hash was calculated during the audit, not recorded at training time. The original data identifier is blank. The task-pinned trainer revision is \texttt{e1d2a730}; both 12B evaluation tasks record evaluator revision \texttt{346798b9} and an empty code diff.\ev{E1--E3}

Zero-shot CSV rows are aligned to the board by recorded order, scene token, category, and label. Their archived question strings are truncated; every recorded question is verified as a prefix of the board question.\ev{E13--E15} This check does not certify historical frame bytes or endpoint precision. In particular, an API-compatible backend name routed to a loopback endpoint must not be treated as proof of hosted BF16 inference.

Thinking reconstruction uses the archive's round-robin journals and ordered failed-item retries.\ev{E8} The last retry-index list is recovered from historical local orchestration and checked item by item against scene token, category, and ground truth. Successful predictions supersede failed attempts. Its checksum and mapping rule are preserved in the shared analysis output. Because the archive task stores rather than executes these runs, its presence alone does not supply full checkpoint fingerprints.

\section{Training History}\label{app:training}
\begin{figure}[ht]\centering\includegraphics[width=\linewidth]{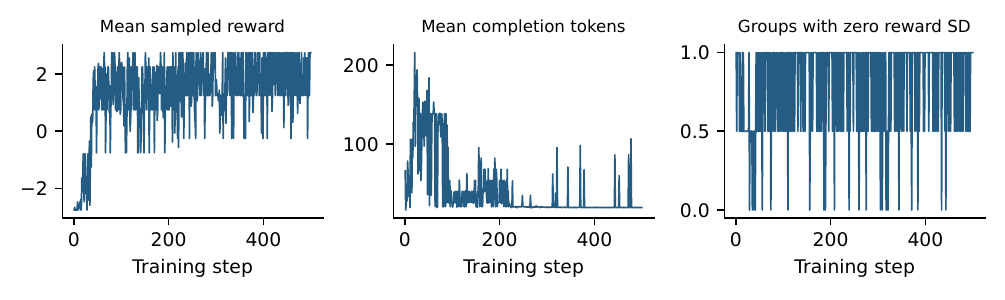}
\caption{Step-indexed Gemma-12B sampled-training traces.\ev{E3} Within-group homogeneous rewards may occur at poor or high reward and do not imply an architectural ceiling.}\label{fig:training}\end{figure}
We use step-indexed scalar history rather than asynchronous latest-value summaries. The terminal group-relative signal observation is confined to the recorded sampled rollouts; no unseen-question convergence result follows from it.

\end{document}